\documentclass{article}

\usepackage{arxiv}
\usepackage{multirow}
\usepackage[utf8]{inputenc} 
\usepackage[T1]{fontenc}    
\usepackage{booktabs}       
\usepackage{nicefrac}       
\usepackage{microtype}      
\usepackage{hyperref}
\usepackage{natbib}
\usepackage{wrapfig}
\usepackage{algorithm2e}
\usepackage{graphicx}
\usepackage{amsmath}
\usepackage{amssymb}
\usepackage{amsfonts}
\usepackage{amsthm}
\usepackage{xcolor}
\usepackage{url,hyperref,cleveref}
\usepackage{subcaption}
\usepackage{bm}
\usepackage{tikz}
\usetikzlibrary{positioning}
\definecolor{tblue}{RGB}{54,116,215}
\definecolor{tred}{RGB}{255,107,107}
\definecolor{tpurple}{RGB}{154,112,163}

\SetKwComment{Comment}{/* }{ */}
\SetKwInput{KwData}{Input}
\SetKwInput{KwResult}{Output}
\RestyleAlgo{ruled}
\DontPrintSemicolon

\newlength{\commentWidth}
\newcommand{\prts}[1]{ \left( {#1} \right) }

\newcommand{\sqbrkt}[1]{ \left[ {#1} \right] }

\title{Spark: \\Modular Spiking Neural Networks}

\author{
    Mario Franco \\
    School of Systems Science and Industrial Enginnering \\ 
    Binghamton University, USA \\
    \texttt{mfrancomndez@binghamton.edu} \\
    \And
    Carlos Gershenson \\
    School of Systems Science and Industrial Enginnering \\ 
    Binghamton University, USA \\
}
\date{}

\begin{document}

\maketitle

\begin{abstract}
    Nowadays, neural networks act as a synonym for artificial intelligence.
    Present neural network models, although remarkably powerful, are inefficient both in terms of data and energy.
    Several alternative forms of neural networks have been proposed to address some of these problems.
    Specifically, spiking neural networks are suitable for efficient hardware implementations.
    However, effective learning algorithms for spiking networks remain elusive, although it is suspected that effective plasticity mechanisms could alleviate the problem of data efficiency.
    Here, we present a new framework for spiking neural networks --- \href{https://github.com/Nogarx/Spark}{\textit{Spark}}\footnote{\url{https://github.com/Nogarx/Spark}} --- built upon the idea of modular design, from simple components to entire models. 
    The aim of this framework is to provide an efficient and streamlined pipeline for spiking neural networks.
    We showcase this framework by solving the sparse-reward cartpole problem with simple plasticity mechanisms.
    We hope that a framework compatible with traditional ML pipelines may accelerate research in the area, specifically for continuous and unbatched learning, akin to the one animals exhibit. 
\end{abstract}


\section{Introduction}

Current artificial intelligence systems inspired by the brain have shown remarkable capabilities in most domains of human interest. 
However, the predominant flavor of neural networks is significantly inefficient, both in terms of energy and information when compared to natural neural networks (also known as brains) \citep{energy_efficiency, information_efficiency}. 
Several alternative artificial neural models have been proposed, with spiking neural networks (SNNs) being the most popular among them due to the possibility of efficient hardware implementations \citep{high_performance_snn, memory_wall} and the collective hunch that effective plasticity mechanisms may open the door to efficient learning as in natural neural networks \citep{information_efficiency}.

Spiking neural networks remain difficult to train for virtually every task. 
SNNs exhibit non-differentiable dynamics which in turn makes them not amenable to the now ubiquitous data-batching plus backpropagation paradigm; although surrogate gradient approaches have shown some promise \citep{surrogate_gradient}.
Critically, it is not clear whether a batched approach is the best way to learn efficiently using SNNs \citep{surrogate_questions}.
Arguably, animal brains do not seem to learn using batched data\footnote{Note that this is not the same as data aggregation.}. 
Still, animal brains are highly effective and efficient at learning, much more than modern ANNs.

Data-batching plus backpropagation reigns supreme among machine learning (ML) practitioners. It offers practical and theoretical guaranties for regular dataset-based problems. 
Nevertheless, this approach struggles with agentic-based problems. Advances in this domain often come from clever mappings into a dataset-based problem or carefully curated loss functions.
These solutions often work well on simple and static problems, i.e., a board game or simple videogames, but it typically fails when the problem drifts faster than what we can dataset-ify it.
As an example, consider the videogame "Minecraft" for which several competitions have been held and is still used as a benchmark for agent-based algorithms \citep{minerl}. 
ANN-agents heavily falter to perform simple tasks after (the equivalent of) hundreds of hours of training. Still, it is not rare to observe a 4-5 year old kid not only surviving, but thriving within a few hours of playing.

We argue that there is a serious case to be held in favor of continuous or iterative learning.
Nevertheless, iterative workflows are computationally prohibitive.  
In the context of ANNs, RNNs faced a similar problem (iterative pipelines are computationally expensive) and lost the battle against the transformer architecture, despite the fact that modern RNNs have a comparable performance \citep{were_rnn_need}.
Performance is critical for iterative pipelines. 
Similarly, SNN research may have been impacted by the lack of an appropriate toolset within the scope of ML. 
Access to spiking-specific-hardware is not widespread and, when available, may present computational restrictions that may play against good research ideas \citep{brain_like_hardware}. 
Several frameworks capable of simulating SNNs are available for generic computational devices \citep{brian2, neuron_py, nest_py}.
However, most of these frameworks were coined for computational neuroscience workflows, where fidelity is a major concern; we may say that several of these frameworks are emulators rather than simulators.
Some of these frameworks are extremely fast, but only for the task for which they were designed: emulation.
In our experience, their out-of-the-box performance tends to falter when confronted with more common ML pipelines, especially with common agent-based pipelines, where information is generated as the system interacts with the environment.

Additionally, past experience with ANNs has shown that compositionality is a potent concept to accelerate ML research.
In contrast, SNNs tend to be implemented as full models; extractions of single components of interest may lead to a lot of code analysis and rewriting, which increases the likelihood of errors and suboptimal code. 

In order to address these problems, in this work we introduce a new computational framework for spiking neural networks --- \textit{Spark} ---, with the goal of providing a common ground for efficient exploration and research of unbatched and iterative learning.
Then, we run several benchmarks to show that \textit{Spark} is an effective and efficient framework for SNN simulation.
Finally, we use \textit{Spark} to solve the sparse-reward cartpole problem.
To the best of our knowledge, this is the first time a SNN is used to solve this problem without relying on surrogate gradients, evolutionary strategies, or any other optimization procedure other than simple plasticity mechanisms.


\section{Spark}

\begin{wrapfigure}{l}{0.33\hsize}
    \includegraphics[width=\hsize]{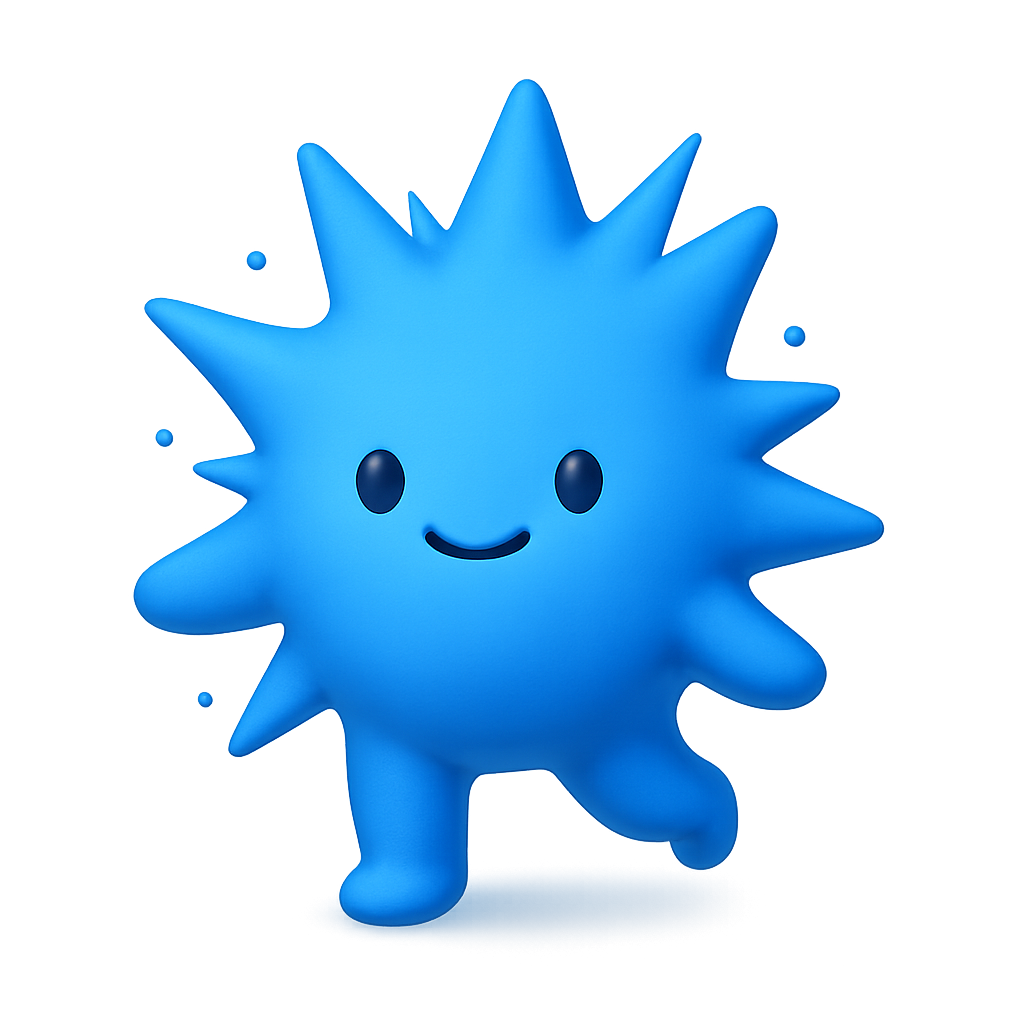}
    \caption{Spark's mascot.}
    \label{fig:mascot}
\end{wrapfigure}

\textit{Spark} is a performant, GPU-based, SNN framework for unbatched iterative learning built on top of JAX for tensor computation and Flax for automatic state management \citep{jax, flax}.
The GPU architecture may not be ideal for SNNs, but SNN's computations are mostly parallel local computations and GPUs are arguably fast for parallel local computation.
More importantly, GPUs are widely available and come with minimum computational restrictions, making them an ideal target platform for a SNN framework. 

At its core, \textit{Spark} decomposes all common computations in a standard SNN pipeline into reusable and performant components.
Every model in \textit{Spark} is a collection of self-contained modules glued together by a minimum amount of code.
By design, we do not impose any strict categorization of what modules or components are.
However, we provide a hopefully useful and reasonable layout for common components, divided into three categories: neuronal components, interfaces and controllers; summarized in table \ref{tab:components}.

Neuronal components are simple, modular descriptions of the usual elements found in regular neuronal models: somas, synapses, delays, plasticity mechanisms, etc.
Most components are implemented in such a way that it is usually possible to replace it with another similar component with minimal to zero effort.

Interfaces implement several (possibly stateful) array mappings and are mostly intended to serve as an I/O interface between the spiking model and the environment.
For example, one common challenge often encountered when using SNNs is the need for "special data", since SNNs only know how to deal with spikes. 
Although this is more a mirage challenge than a real one, it introduces systemic biases through specific mappings performed to the datasets, e.g., time vs rate vs phase codings \citep{rate_vs_temp, neural_coding_review}.
Perhaps, even more problematically, this can lead to unconscious biases that SNNs are special and need special data, especially among new practitioners, neither of which is particularly desirable.
A similar case can be constructed for the end-side of the network.
It is common to use a secondary model to map the output of the SNN, e.g., linear regression, Fisher linear discriminant, etc. \citep{simple_framework_snn, liquid_snn, plasticity_cards}.
In principle, there is nothing wrong with this approach, SNN are viable computational reservoirs, but it misses the point of using SNNs in the first place.
As before, a subconscious bias that SNNs may not be capable of directly outputting solutions to real problems is not desirable. 

Controllers are used to simplify model construction by abstracting models as information flow graphs.
One way to think about them is as small "transpilers" that map a model template or configuration into a generic, yet performant, instantiation of a model.
Under the hood, they simply try to layout a model as efficiently as possible for the JIT compiler, by reorganizing the order of execution of submodules, introducing caches, etc. 
Traditional model definition is still viable and, in some cases, it may even be encourage, but in our internal experience with the framework we found controllers to be an extremely potent tool for model exploration.  

\begin{table}
    \centering
    \begin{tabular}{ccc} 
    \rule{3cm}{0pt}&\rule{3cm}{0pt}&\rule{4cm}{0pt}\\[-\arraystretch\normalbaselineskip]
        \toprule
            \textbf{Category} & \textbf{Type} & \textbf{Examples / Usage} \\ 
        \midrule
        \multirow{4.75}{*}{\shortstack{Neuronal \\ Components}}
            & Soma & LIF model, AdEx model \\ 
            \cmidrule(l){2-3}
            & Synapses & Linear synapses, traced synapses \\ 
            \cmidrule(l){2-3}
            & Learning Rules & Hebbian rule, Oja's rule \\
            \cmidrule(l){2-3}
            & Delays & Neuron-to-neuron delays \\
        \cmidrule(l){1-3}
        \multirow{3.5}{*}{Interfaces} 
            & Input & Numeric values to spike streams \\ 
            \cmidrule{2-3} 
            & Output & Spike streams to numeric values \\ 
            \cmidrule{2-3} 
            & Control & Reshape, subsampling \\
        \cmidrule(l){1-3}
        \multirow{2.25}{*}{Controllers} 
            & Neurons & Arbitrary neuronal model constructor \\ 
            \cmidrule{2-3} 
            & Brains & Arbitrary model constructor \\ 
        \bottomrule
    \end{tabular}
    \vspace{2mm}
    \caption{
        \label{tab:components} 
        Default component convention used in \textit{Spark}.
        Note that this convention is not strict but it provides a useful set of programming templates.
    }
\end{table}

\begin{figure}
        \centering
        \includegraphics[width=1\hsize]{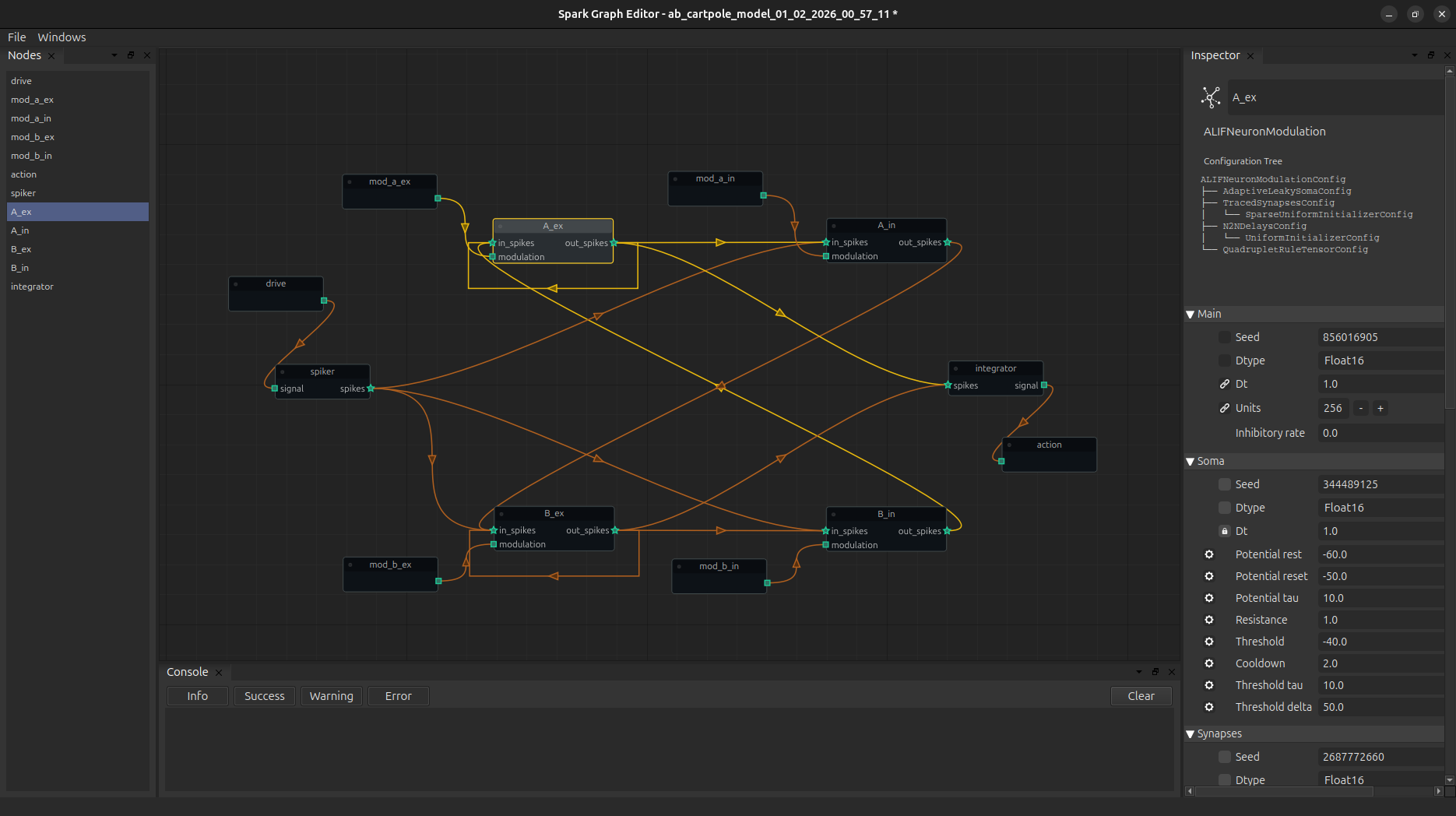}
        \caption{
            A glimpse into Spark's graphical interface. 
            Source nodes represent inputs to the model, while sink nodes represent outputs.
            The graph editor allows for simple and quick model design. 
            Currently not all features of \textit{Spark} are supported through the editor but models can be imported/exported to allow for blueprint edition via code as necessary.
        }
        \label{fig:gui}
\end{figure}

Building models with code is the standard approach in any machine learning library.
This approach works terrifically well for ANN, which consists mostly of stacks of some common building blocks.
However, building more intricate architectures can prove extremely challenging when one wanders away from this simple pattern.
Moreover, we speculate that architectural biases are one of the major missing ingredients to make SNNs work as well as any natural neural network.
For this reason, \textit{Spark} is also bundled with a lightweight graphical editor (figure \ref{fig:gui}).
This graphical editor opens the possibility to design and configure complex models without zero code involved.
Such models can be later exported and edited with code when necessary, minimizing design time and errors, while maintaining the flexibility of systemic editions.

\textit{Spark} abstracts the blueprint of a model from the executable model itself. 
This greatly improves the reproducibility and shareability of the models among users; any model template gets condensed into a single file that can be later loaded, modified or expanded at ease, either via code or the GUI. 
We consider this feature to be critical to accelerate SNNs research.
It is common for works on SNNs to only include formal mathematical details and, when code is provided, it is not rare that extracting or extending the code becomes a challenge on its own (everybody programs for themselves and rightly so).
However, when sharing code, it is important to minimize these frictions as much as possible; the blueprint system plus the graphical editor is a palliative to this problem.


\section{Benchmarks}


\subsection{Fidelity}

\begin{figure}
        \centering
        \includegraphics[width=0.9\linewidth]{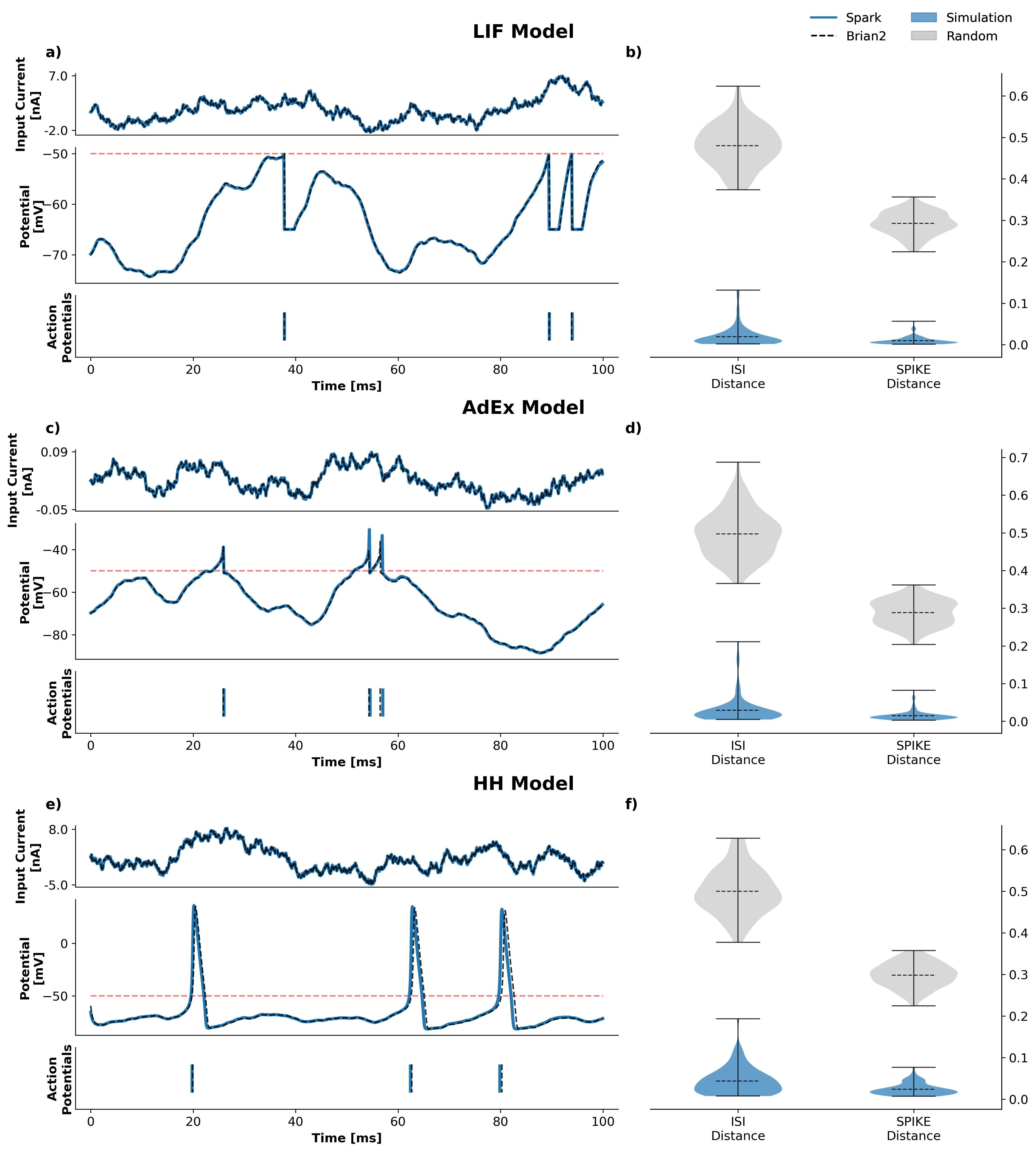}
        \caption{
            Fidelity benchmark. 
            Three soma simulations of increasing complexity, LIF (a,b), AdEX (c,d) and HH (e,f) are used to showcase the fidelity of \textit{Spark} simulations. 
            As a reference for comparison we use Brian2 and model implementations used in the official Brian2 documentation.
            Simulations are driven by an Ornstein–Uhlenbeck process-like current for one second.
            Left subplots (a, c, e) show the input current (top subplot), the membrane potential (middle subplot) and the output spikes (bottom subplot) for a single trial of 100 ms.
            Right subplots (b, d, f) show two common train spike statistics: the average ISI distance and the average SPIKE-distance for 100 simulations, one second each. 
            For comparison, we use a shuffle version of the simulation spike trains. 
        }
        \label{fig:fidelity}
\end{figure}

Although \textit{Spark} is not explicitly designed for precise emulation of neuronal dynamics, it is capable of replicating standard differential-equation-based models with good fidelity, even in low precision mode (float16).
Some proposed learning mechanisms are based on exact spike timing or membrane potentials.
As such, simulation fidelity is important to properly implement such mechanisms.

In order to compare the fidelity of \textit{Spark}, we use Brian2 \citep{brian2} a popular, efficient and well established framework in computational neuroscience.
For this benchmark, we limit the comparison to the dynamics of the soma and use a common selection of models: a Leaky Integrate-and-Fire (LIF) model, an Adaptive Exponential (AdEx) model and a Hodgkin-Huxley (HH) model. 
Soma dynamics are often the most complex part of any neuronal model and are a good proxy for the quality of the simulation.

One hundred simulations were conducted using a deltatime $\Delta t = 0.1\;ms$ for the LIF and AdEx models and $\Delta t = 0.05\;ms$ for the HH model. 
Neurons are driven by an Ornstein–Uhlenbeck process-like current for one second.
In order to compare the two simulations, we use the ISI-distance and the SPIKE-distance \citep{isi_distance, spike_distance}.
The distances of the shuffle train spikes are used as a reference.

For Brian2 simulations, we always use the best differential equation solver available and use standard precision (32-bits).
\textit{Spark} does not rely on any explicit solver.
Throughout the entire framework, we integrate most differential equations with the standard Euler method, falling back to the exponential Euler method when the standard method has been observed to be unstable or unaccurate.
Since \textit{Spark} is all about performance, its default execution mode is low precision.
Consequently, we ran all models in this benchmark using low precision floats (16-bits) only.

The benchmark results are summarized in figure \ref{fig:fidelity}.
Train spike distances between \textit{Spark} and Brian2 are closely distributed around and skewed towards zero, which indicates a good agreement between both simulations.
Moreover, visual inspection of the soma potential dynamics suggests that both simulations are in good agreement.
Further visual inspection shows that, even when one model, either Brian2 or Spark, produces an extra spike (or fails to produce one), both simulations tend to quickly synchronize again (data not shown).

Numerical integration using low precision floats introduces other challenges that are important to consider: operations must be constrained to specific ranges, the order of the operations may become critical for stable computation, certain functions like exponentials may become unstable, etc.
Low precision offers clear speed advantages but it is still possible to opt for standard precision when and where necessary.
This opens the door to hybrid-precision computation: low precision for simple and fast integration or high precision for slow but accurate integration, depending on the needs of each particular component.
It is important to mention that, currently, this is not a standard feature.
Common scenarios are expected to work out-of-the-box, although some cases may require manual broadcasting in order to prevent the entire model from falling back to standard precision.


\subsection{Performance}

\begin{figure}
    \centering
    \includegraphics[width=1\linewidth]{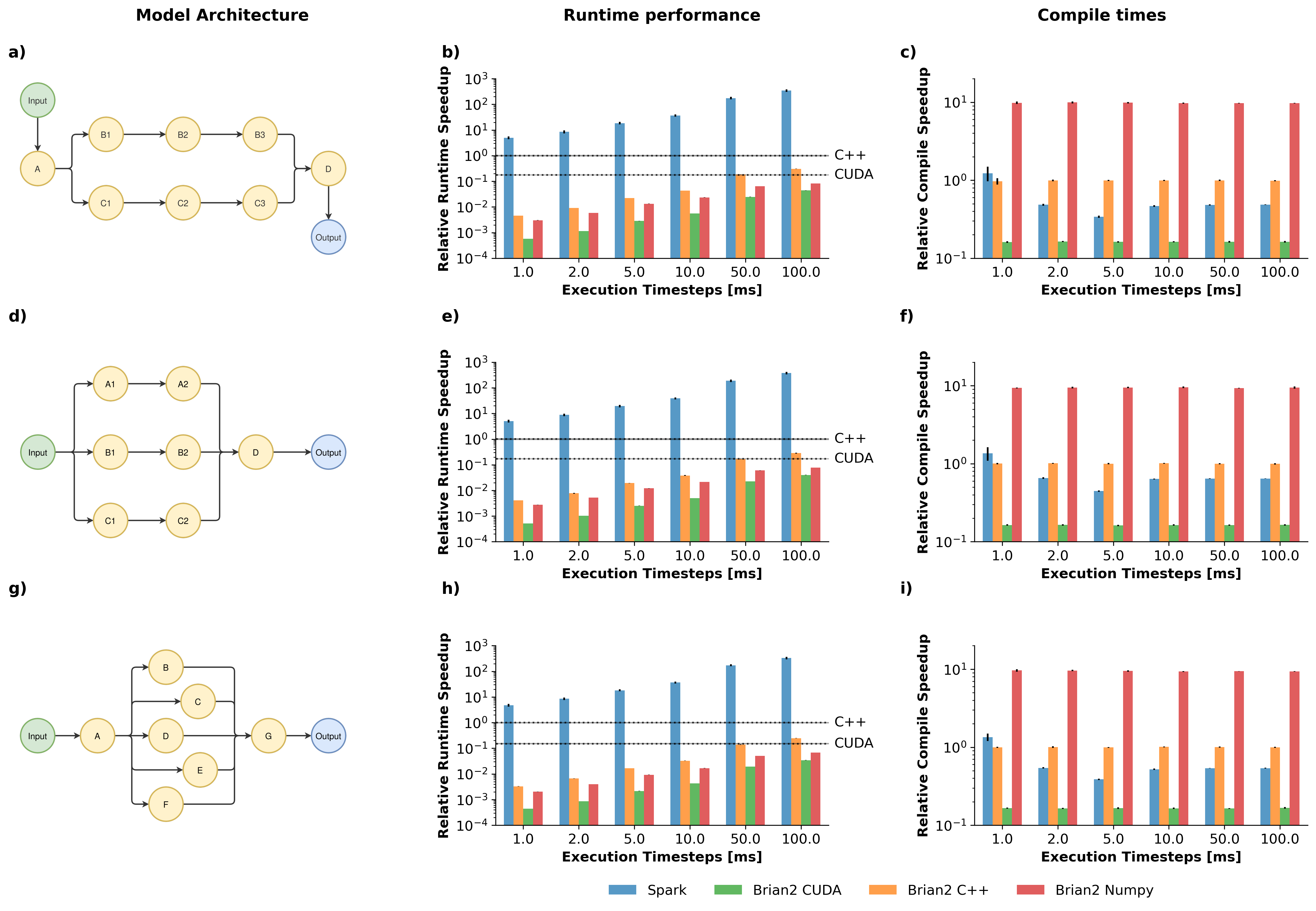}
    \caption{
        Performance benchmark of three different network architectures. 
        (a,d,g) Network architectures used for the simulations. Each node represents a "pool of neurons" of 1024 units, 20\% of which are inhibitory neurons. 
        (b,e,h) Average relative speedup of 10 seconds simulations, 25 network instantiations, 5 runs per instantiation, for architectures in (a,d,g), respectively.
        Speedups are normalized with respect to a C++ single execution (higher is faster).
        Execution timesteps indicate how many timesteps each model computes before a new input signal is presented to the network and a new output signal is registered from the network. 
        Dashed lines denote the best-case scenario, i.e., when the network is not interrupted, for Brian2 with the C++ and CUDA backends.
        (c,f,i) Average compile times for each instantiated network model, for architectures in (a,d,g), respectively.
        Compilation is considered as the time for object construction plus the C++/JIT compilation required to run the network; numpy is considered as the time to instantiate the network object.
    }
    \label{fig:performance}
\end{figure}

Modern computational hardware is remarkably powerful but it is not compatible with SNN at a fundamental level. 
Almost any implementation of a SNN requires solving a series of differential equations, chained in a temporal sequence to produce a single answer.
The problem becomes even more complicated in an iterative learning paradigm; in such a context it is not possible to run multiple instances of the same network and "average" their responses.

Moreover, when considering an iterative learning paradigm, performance becomes critical; how much one can compute directly links to how much one can learn.
For this reason, \textit{Spark} puts special attention on enabling an interactive and performant simulation. 

In order to showcase the capabilities of \textit{Spark}, we again use Brian2 as a reference; Brian2 has comparable or better performance than other similar frameworks \citep{brian2}. 
Although Brian2 was designed for accurate simulations, Brian2's C++ compiled models are extremely performant. 

Our testbed consists of three different network architectures, 25 network instantiations per architecture and 5 runs per instantiation for a span of 10 seconds.
Simulations are given a timeout of 60 seconds, after which, simulation time is estimated based on the number of iterations completed.
Simulations were conducted assuming different interaction times: $1ms$, $2ms$, $5ms$, $10ms$, $50ms$ and $100ms$. 
An interaction step indicates how much time each model computes before a new input signal is presented to the network and a new output signal is read from the network.
Write-in and read-out operations are expensive in CPU's and even more so in GPU's.
A good balance of write-in and read-out is necessary to get the most out of a SNN.
Note that use of the concepts of "interaction times" and "$ms$" here may be an overstretch, and they are just a bioinspired abstraction for how much computation do we packed between I/O operations.

It is important to remark that interactive simulations are not a main feature of Brian2 and it is only possible to do truly interactive simulations while using a numpy backend (low performant) or by writing specialized C++ code to interact with the environment.
To keep our implementation as simple as possible, we do interaction "mock ups" when using Brian2. 
Brian2 models are compiled with the input embedded into the network itself, then the models are executed $k$ times, for the interaction times $t$ indicated above, for a total time $T_{max}$, such that a single run is split into $k$ smaller runs.
Additionally, we add a reference time of running Brian2 without stopping in any way, as a reference for the minimum time possible achievable by Brian2. 

Brian2 also offers GPU implementations \citep{brian2cuda, brian2genn}, although only the CUDA implementation is actively maintained.
Moreover, our benchmark network architectures involve a substantial number of spike propagation operations which are known to significantly slow down the CUDA backend \citep{brian2cuda}; we tested several backend configurations using one of our network architectures and ran all simulations with the most performant setup. 

Runtimes and compilation times are shown in figure \ref{fig:performance}.
As expected, the performance of all the simulations increases as the interaction time increases: stopping a simulation to do an interactive step is a costly operation, even when no cross-communication between the CPU and the GPU is necessary.
\textit{Spark} significantly outperforms Brian2 implementations, regardless of the backend.
Our framework is up to $\sim5$ to $\sim350$ times faster, depending on the interaction steps, than running a single execution of Brian2 C++ on our testbed.
Major speedups, of up to 3 orders of magnitude, are observed when compared to the rest of the interaction mock-up simulations.
Brian2 CUDA was slower than all CPU-based implementations in our testbed.
According to \cite{brian2cuda}, our network size ($\sim10^{4}$ units per pool) may be too small for Brian2CUDA to take advantage and, as mentioned above, spike propagation operations are quite abundant in our simulations.
This operation severely impacts the CUDA backend performance and can even make it slower than the C++ backend.
Note that Brian2 offers several advanced features for direct manipulation of the compiled code which, in theory, may be used to increase the performance for interactive simulations.
However, such modifications may require a deep understanding of Brian2's implementation, making them unavailable to most SNN practitioners. 

\textit{Spark} is slightly slower at compile time.
Moreover, compilation time of a \textit{Spark} model is not fixed; times may increase significantly when more than one execution mode is required (e.g., plain execution, reward delivery, spike recording, etc.).
Compilation times increase linearly with the number of execution modes required, approximately.
Basic understanding of Jax's JIT compiler is required to avoid unnecessary compilations.
In contrast, Brian2 offers fixed compilation times, regardless of the number of units in the network and the execution time.
Nonetheless, \textit{Spark} compilation times are comparable to Brian2 compilation times.

Overall, we can observe that \textit{Spark} is significantly faster, for interactive simulations, than Brian2, while maintaining similar compilation times.


\section{Cartpole: a simple study case}

The Cartpole problem is a classical control problem where the goal is to balance a rigid pole that is mounted on top of a cart by moving the cart either to the left or to the right.
Today, this problem may be considered as a rite of passage for anyone interested in autonomous agents or reinforcement learning and is trivially solved in classrooms all around the world by means of deep learning.
Nonetheless, SNNs still struggle to solve it and it is still considered a benchmark for SNN \cite{cartpole_as_benchmark}.
Previous works used surrogate gradients \citep{snn_cartpole_surrogate}, evolutionary strategies \citep{snn_cartpole_evolve}, actor-critic architecture \citep{snn_cartpole_reward_mod} and reward-modulated synaptic plasticity \citep{snn_cartpole_stdp_actor_critic}; all with a mixed degree of success.  

We showcase \textit{Spark} by addressing and solving this classic control problem.
Our approach to the carpole problem consists of a reasonable architecture bias, a three-factor modulated plasticity rule and letting the network do what it whats to do (self-organize) with some sporadic feedback.

In the brain, neurons do not "randomly" form connections among each other.
Examples of biased connectivity can be observed throughout the entire brain, such as the V1 visual cortex and the CA3 hippocampal area \citep{principles_neural_design}, just to mention a few.
Here we take seriously that this bias matters.
Thus, we consider a model of two populations that are architecturally biased to mutually inhibit each other. 
For the sake of exposition, we refer to these populations as "left" and "right", depending on the action they trigger.

\begin{figure}
    \centering
        \includegraphics[width=0.9\linewidth]{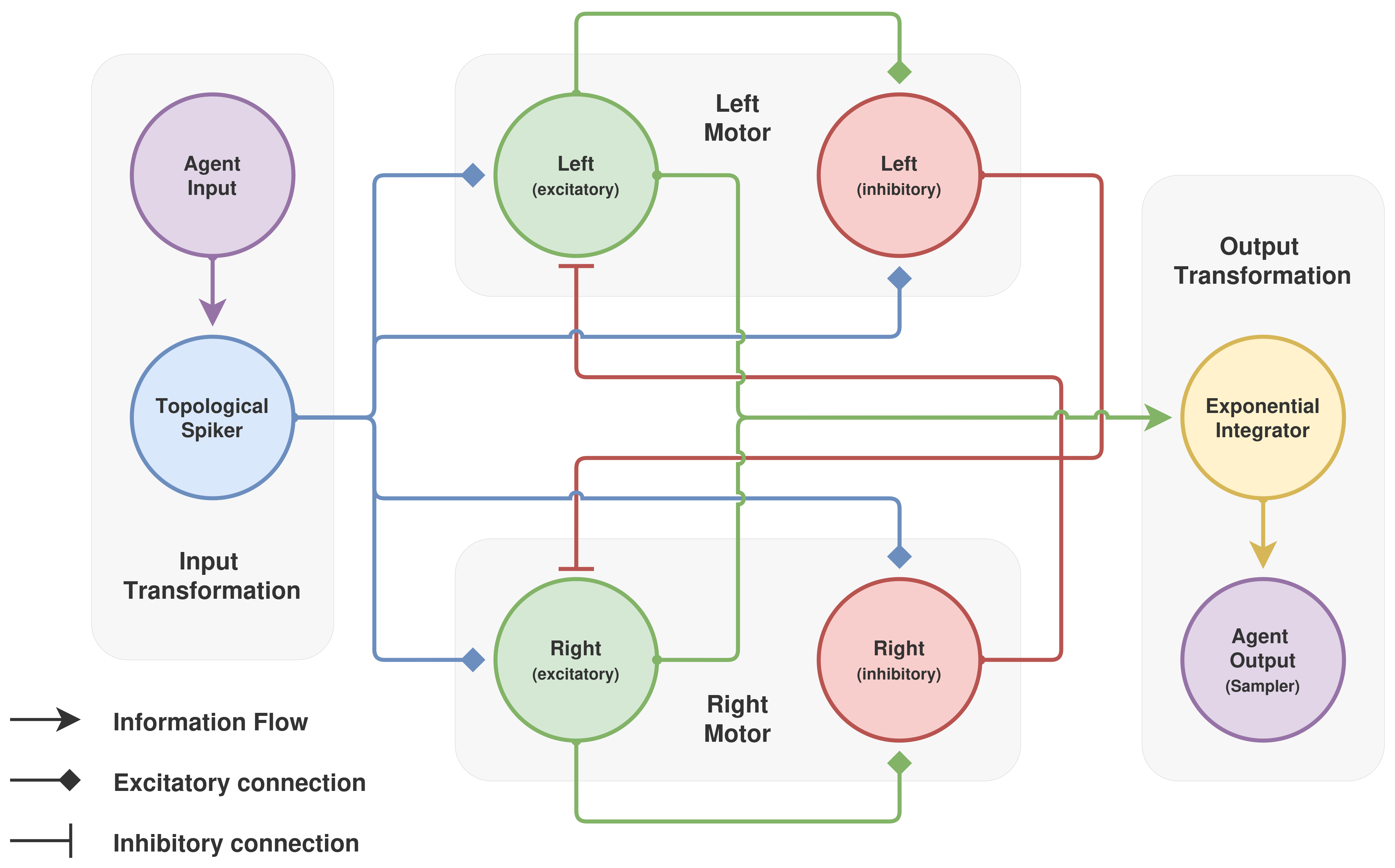}
        \caption{
            Cartpole model diagram.
            The model consists of a simple A vs B architecture; activity on A inhibits activity on B and vice versa.
            Agent input is passed through a topological Spiker which maps single values (e.g. an angle) into a distribution of spikes.
            Agent output consists of a traced "measurement" of the activity of each excitatory subpopulation.
        }
        \label{fig:model_diagram}
\end{figure}

To complete the model we need to define the I/O interface.
Within \textit{Spark}, this is accomplished by means of two types of modules that we named "spikers" and "integrators" for input and output, respectively.

At the input side of the model we use a Topological Spiker, which can be thought as a grid-cell-like mapping. 
Inputs with certain topological properties are mapped to distributions of spikes that preserve such topological features by appropriately gluing the borders of the input data.
Note that this operation is not automatic and requires prior knowledge of the data; for example, a linear variable should be mapped into a line and an angular variable should be mapped into a circle.
Given the dynamics of the observations of the cartpole environment, in our model everything is mapped to line segments.

The output is handled with a simple exponential integrator.
This integrator is a simple collection of $k$ saturable traces, which we feed using the spiking activity of the "left" and "right" populations.
In our model, $k=2$ and each trace is exclusively fed using "left" or "right" spikes. 
We adopt a greedy policy and choose the associated action of the largest trace at the moment of the interaction timestep.
For implementation details about the I/O interfaces we refer the reader to the code repository.
A summary of the model architecture can be seen in figure \ref{fig:model_diagram}.

All neurons in the model are implemented as Leaky Integrate-and-Fire (LIF) neurons  \citep{neuronal_dynamics_gerstner} with a refractory period and threshold adaptation.
Excitatory and inhibitory nodes consist of 256 and 64 neurons each, respectively.
Synapses are implemented as simple traced currents with neuron-to-neuron specific delays.
Weights are uniformly and sparsely initialized. 
For the sake of completion we summarize the neuronal model with the following set of equations,
\begin{align}
    \tau_{m} \frac{dv_{i}\prts{t}}{dt} &= 
        \begin{cases} 
            \;-\prts{v_{i}\prts{t} - v_{rest}} + R \sum I_{ij} \prts{t} &\text{if} \quad r_{i}(t) > 0\\
            \;\qquad\quad -\prts{v_{i}\prts{t} - v_{rest}} &\text{ otherwise}
        \end{cases} \\
    \tau_{th} \frac{d\theta_{i}\prts{t}}{dt} &= -\prts{\theta_{i}\prts{t} - \theta_{base}} + a S_{i} \prts{t} \\
    S_{i}\prts{t} &= 
        \begin{cases} 
            \;1 &\text{if} \quad v_{i}\prts{t} > \theta_{i}\prts{t}\\
            \;0 &\text{ otherwise}
        \end{cases} \\
    \frac{dr_{i}\prts{t}}{dt} &= 
        \begin{cases} 
            \;-r_{spike} &\text{if} \quad S_{i}\prts{t} = 1\\
            \;\,dt &\text{ otherwise}
        \end{cases} \\
    \tau_{s} \frac{dI_{ij}\prts{t}}{dt} &= -I_{ij}\prts{t} - w_{ij} S_{i}\prts{t - d_{ij}}
\end{align}
Where $v_{i}$ denotes the membrame potential of the $i-th$ neuron, $\tau_{m}$ is the membrane time constant, $v_{rest}$ is the membrane rest potential, $R$ is the membrane resistance, $I_{ij}$ is the synaptic current between neurons $i$ and $j$, $r_{i}$ is a refractory variable, $\theta_{i}$ is the action potential threshold of the $i-th$ neuron, $\tau_{th}$ the threshold time constant $\theta_{base}$ is the base action potential threshold, $a$ is the threshold increment per spike, $S_{i}$ indicates wheter the $i-th$ neuron spiked or not, $r_{spike}$ is the refractory time, $w_{ij}$ is the synaptic strength between neurons $i$ and $j$ and $d_{ij}$ is the time delay between neurons $i$ and $j$. Equations are solved with a mixture of the euler forward and the exponential euler methods using a $\Delta t = 1\;ms$.

For the plasticity mechanism, we use a simple three-factor quadruplet (STDP) plasticity rule \citep{3_factor_rule}, given by,
\begin{align}
    \frac{dw_{ij}}{dt} &= \eta M_{3rd}\prts{t} \prts{
        S_{i}\prts{t}\sqbrkt{\alpha_{ij} + \beta_{ij} x_{ij}^{pre}\prts{t}} +
        S_{j}\prts{t - d_{ij}}\sqbrkt{\gamma_{ij} + \delta_{ij} x_{ji}^{post}\prts{t}}
    }
\end{align}
Where $\eta$ is the learning rate, $M_{3rd}$ is a modulatory third factor, $x_{ij}^{pre}$ and $ x_{ji}^{post}$ are the pre and post synaptic eligibility traces, respectively, and $\alpha_{ij}$, $\beta_{ij}$, $\gamma_{ij}$ and $\delta_{ij}$ are connection-dependent scaling constants.
We set the value of $\alpha_{ij}$, $\beta_{ij}$, $\gamma_{ij}$ and $\delta_{ij}$ according to \cite{plasticity_cards} for a stable dynamic that favors sequential tasks.

The modulatory factor is computed as a combination of an exponential moving average of the number of steps the agent performs each episode, the episode reward and an exponential decay, 
\begin{align}
    R_{step} &= 
        \begin{cases} 
            \;-1 &\text{if pole fell} \\
            \;-1 &\text{if agent oob} \\
            \;\,0 &\text{otherwise}
        \end{cases} \\
    M_{3rd}\prts{t} &= 
        \begin{cases} 
            \;\sqrt{\frac{N_{ema}}{N_{max}}}R_{step}\exp\prts{-\tau_{R}t} &\text{if} \quad R_{step} \neq 0 \\
            \;\,\lambda &\text{ otherwise}
        \end{cases}
\end{align}
Where $N_{ema}$ is the exponential moving average of the number of steps, $N_{max}$ is the maximum number of steps in the cartpole environment $\prts{500}$ and $\tau_{R}$ is the time constant of the modulatory signal. 
It is important to mention that we deliver $M_{3rd}\prts{t}$ independently to each motor, where the action that we want to suppress receives  $M_{3rd}\prts{t}$.
Additionally, we found that feeding  $-\xi M_{3rd}\prts{t}$, with $\xi$ a small positive constant to the other motor, helps stabilize the agent since it slightly promoted synaptic growth.
Note that, the first term $\prts{\sqrt{\frac{N_{ema}}{N_{max}}}}$ is not strictly necessary, it is still possible to successfully train agents without it.
However, this term significantly helps stabilize late agent behavior, as it tends to prevent small agent mistakes from driving the learning process.

\begin{figure}
    \centering
    \includegraphics[width=1\linewidth]{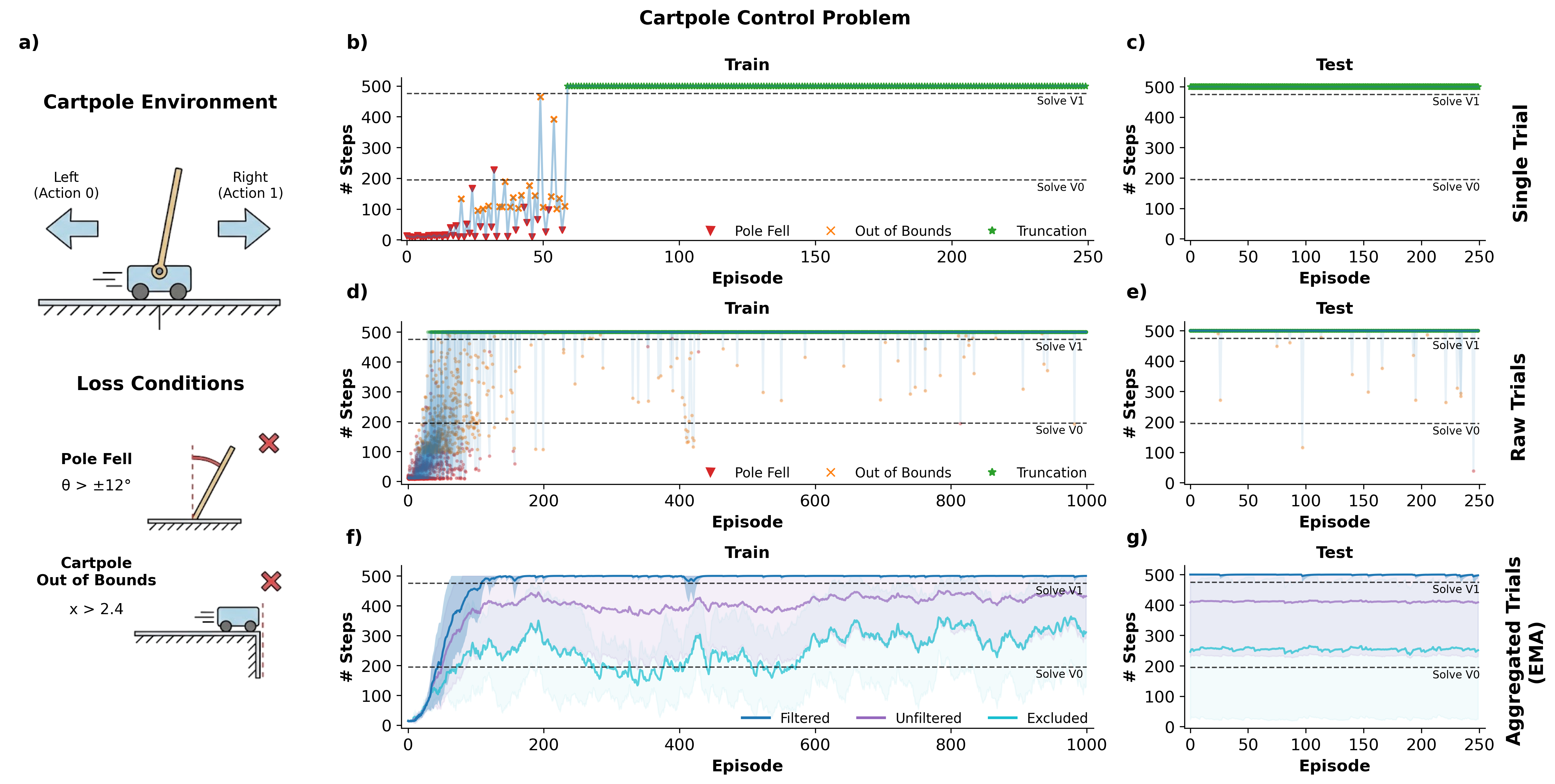}
    \caption{
        Cartpole Control Problem.
        (a) Summary of the cartpole environment. 
        Agents need to balance a pole that is mounted on top of a cart by pushing the cart to the left or right, without moving the cart too far from the central position.
        (b) Training trajectories of a single agent, with episode end annotations.   
        (c) Test performance for the agent (b), after 1000 episodes.
        (d) Training trajectories of 16 agents for 1000 episodes.
        (e) Test performance for the same 16 agents.
        (f,g) Average of the exponential moving average ($\tau=0.8$) of performance of 25 agents.
        Filtered denotes the EMA for the 16 agents of subplots (d,e).
        Agents under the filtered category were capable of solving the cartpole problem within $\sim$40-80 episodes.
        Unfiltered denotes the EMA for all 25 trained agents. 
        Excluded denotes the EMA of the 9 agents that failed to stabilize quickly.
        The extra 9 agents in unfiltered encompasses agents for which no stabilization was observed in 1000 episodes (6) and agents that took more than 500 episodes to stabilize (3).
        Note that all 9 excluded agents still performed better than random and may still solve the version zero of the environment (200 steps).
    }
    \label{fig:cartpole}
\end{figure}

Networks are trained in an online setting, i.e., we update the network as it interacts with the environment, using an interaction time of $\Delta t = 50\;ms$, i.e., we fix the input of the network for $50\;ms$ then we read the output and set a new input.
Moreover, we adopted a sparse reward scenario, networks are given a single reward signal at the end of the episode. 
For the rest of the episode, we let the network do what it wants to do (self-organize) by setting the modulatory factor to a fixed, relatively small, constant.
When the episode terminates, we run the network for another extra step, using the final observation.
During this extra iteration, we send an exponentially decaying but strong modulatory signal, conditioned on the finish state triggered by the environment. 
For specific implementation details we refer the reader to the paper \href{https://github.com/Nogarx/Spark_paper}{repository}\footnote{\url{https://github.com/Nogarx/Spark_paper}}.

We ran 25 simulations of 1000 episodes each using the approach described above.
After 1000 episodes, we test the network for an additional 250 episodes. 
Although one of the major motivations of this work is to design systems that can learn continuously, during testing, we freeze the network, preventing any further adaptation.
Figure \ref{fig:cartpole} shows the simulation results.

From figure \ref{fig:cartpole} we can see that SNNs agents are capable of solving the cartpole control problem remarkably fast.
16 of the 25 agents were able to obtain a perfect score  and stabilize within 40 to 80 episodes, approximately.
Three of the remaining nine agents stabilized after more than 500 episodes, while no stabilization was observed for the other six agents. 
Nonetheless, these nine agents performed above random, usually above the "solve v0" performance, which corresponds to the minimum number of steps required to solve the easy version of the cartpole problem. 
Manual inspection of the unstable models suggests that these models struggle with boundary conditions of the environment rather than having trouble stabilizing the pole.
We suspect that stabilization was a matter of time for those 6 agents since the plasticity mechanism supports the addition and deletion of connections.
Moreover, better initialization priors for the synapses could drastically improve the performance, since it is the only difference among any two agents.

For this particular problem and model the final behaviour exhibit a quite intuitive policy: if the pole is balanced any action is good; if the pole starts to fall to either side, the "left" or "right" populations quickly take control of the cart basically silencing the other population; and if the cart starts to go out of bounds, a similarly mechanisms kicks in quickly returning the cart back to the stage. 
For a small showcase of these models in action, we refer the reader to the supplementary \href{https://www.youtube.com/playlist?list=PLtT9354ygtrvI0mpktdr6OCqbirBc_avV}{ videos}\footnote{\url{https://www.youtube.com/playlist?list=PLtT9354ygtrvI0mpktdr6OCqbirBc_avV}}.


\section{Discussion}

We introduce a flexible and extendable GPU-based framework --- \textit{Spark} --- for building and running modular spiking neural networks.
\textit{Spark} offers a great balance between simulation performance and fidelity.
Moreover, \textit{Spark} is bundled with several utilities to streamline and simplify the entire pipeline of design, construction and execution SNNs; I/O interfaces, controllers and graphical interfaces, etc.
Furthermore, a dual representation of the models as blueprint-instance, drastically increases the reproducibility of any \textit{Spark} model and provides a great starting point for future research ideas since model blueprints are easy to share and modify/extend later on. 

We show how \textit{Spark} can be used to easily solve the cartpole control problem, currently considered a benchmark for SNNs \citep{cartpole_as_benchmark}. 
Several SNN models have been proposed to solve this problem, which have several overlaps with our solution \citep{snn_cartpole_evolve, snn_cartpole_surrogate, snn_cartpole_stdp_actor_critic, snn_cartpole_reward_mod}. 
Our approach relies on a simple architectural bias and modulated plasticity rules and can define an I/O interface which allows for direct interaction with the environment.
We argue that this combination of features allowed us to train models capable of solving the cartpole problem extremely fast (40 to 80 episodes), approximately.

In contrast, standard deep reinforcement learning algorithms (e.g, DQNs, SARSA, REINFORCE, etc...) may take up to 500-1000 episodes to obtain a good solution.
Even modern implementations take around 100 to 200 episodes, approximately, to obtain a model capable of solving the task consistently \citep{cartpole_nfq2, cartpole_tutorial_torch}. 
We suspect that our algorithm is the current SOTA (for sample efficiency) on this problem, but no leaderboard is maintained for this classic problem. 
To the best of our knowledge, this is the first time a SNN is able to properly solve this problem without relying on surrogate gradients, evolutionary algorithms or other optimization techniques. 
However, our algorithm is not perfect.
Although most of the time our approach obtains a solution extremely fast, sometimes it fails to stabilize quickly and in some cases no stabilization was observed during the 1000 episode limit. 
Moreover, the modulatory factor $M_{3rd}$ is still handcrafted, which limits the scope of this approach.
We speculate that a simple critic network may be used to streamline the computation of this factor.

Finally, we want to mention that \textit{Spark} is still under active construction and we expect it to continue growing and getting better over time.
Model selection is currently limited, but we are actively working on adding the most common models that can be found in the literature as building blocks.
Similarly, there is ample room for optimization since currently most operations are implemented as standard jax operations; custom kernels tailored for the job may provide significant speeds up through the entire ecosystem.
Neuroscience is rich in tools for studying and understanding the brain; most of these techniques can be reimagined/translated for SNNs, providing a starting ground for interpretability studies on SNNs. 


\section{Materials}
Code for the package can be accessed through \textit{Spark} \href{https://github.com/Nogarx/Spark}{ repository}\footnote{\url{https://github.com/Nogarx/Spark}}.
Code for the simulations and figures for the present work can be accessed through \textit{Spark} \href{https://github.com/Nogarx/Spark_paper}{paper}\footnote{\url{https://github.com/Nogarx/Spark_paper}}


\section{Acknowledgments}

The authors acknowledge Dr. Chubs for fruitful discussions and encouragement.


\bibliographystyle{abbrv}
\bibliography{references}


\end{document}